\documentclass{article}

\usepackage[nonatbib,final]{nips_2018}

\usepackage[utf8]{inputenc} 
\usepackage[T1]{fontenc}    
\usepackage{hyperref}       
\usepackage{url}            
\usepackage{booktabs}       
\usepackage{amsfonts}       
\usepackage{nicefrac}       
\usepackage{microtype}      

\usepackage{amsmath,mathtools,dsfont,lipsum}
\usepackage{amsfonts}
\usepackage[toc]{appendix}
\usepackage{subcaption}
\usepackage{nth}
\usepackage{verbatim}
\usepackage{hyperref}
\usepackage{fancybox,framed}
\usepackage{float}
\usepackage{multirow}
\usepackage{color}
\usepackage{amssymb}
\usepackage[table]{xcolor}
\usepackage[usestackEOL]{stackengine}
\DeclarePairedDelimiterX{\infdivx}[2]{(}{)}{%
  #1\;\delimsize\|\;#2%
}

\newcommand{\rpm}{\raisebox{.2ex}{$\scriptstyle\pm$}}

\begin{document}

\title{Mixture Density Generative Adversarial Networks}
\author{
  Hamid Eghbal-zadeh$^{1}$
  \thanks{This work was supported by the Austrian Ministry for Transport, Innovation \& Technology, the Ministry of Science, Research \& Economy, and the Province of Upper Austria in the frame of the COMET center SCCH.}
  \ \ \ \ \ \ \ \ \ \ \ 
  Werner Zellinger$^2$
  \ \ \ \ \ \ \ \ \ \ \ 
  Gerhard Widmer$^{1}$  \\ \\
  $^1$~LIT AI Lab \& Institute of Computational Perception\\
  $^2$~Department of Knowledge-Based Mathematical Systems\\
  Johannes Kepler University of Linz, Austria\\
  \texttt{\{hamid.eghbal-zadeh, werner.zellinger, gerhard.widmer\}@jku.at}\\
}

\maketitle
\begin{abstract}
Generative Adversarial Networks have surprising ability for generating sharp and realistic images, though they are known to suffer from the so-called \emph{mode collapse problem}.
In this paper, we propose a new GAN variant called Mixture Density GAN that while being capable of generating high-quality images, overcomes this problem by encouraging the Discriminator to form clusters in its embedding space, which in turn leads the Generator to exploit these and discover 
different modes in the data.
This is achieved by positioning Gaussian density
functions in the corners of a simplex, using the resulting
Gaussian mixture as a likelihood function over discriminator
embeddings, and formulating an objective function for GAN training that is based on these likelihoods.
We demonstrate empirically (1) the quality of the generated images in Mixture Density GAN and their strong similarity to real images, as measured by the
Fr\'echet Inception Distance (FID), which compares very favourably with state-of-the-art methods, and (2) the ability to avoid mode collapse and discover all data modes.
\end{abstract}

\section{Introduction}
\label{sec:intro}

Generative Adversarial Networks (GANs)~\cite{goodfellow2014generative} learn an implicit estimate of the Probability Density Function (PDF) underlying a set of training data and can learn to generate realistic new samples.
One of the known issues in GANs is the so-called \emph{mode collapse}~\cite{arjovsky2017towards,goodfellow2016nips,mescheder2017numerics}, where the generator memorizes a few training examples and all the generated examples are similar to only those. 
These memorized examples are known as \emph{modes}.
Although for the generator having a few -- but good -- modes for generated images is enough to fool the discriminator, 
the result is not a good generative model:
when mode collapse happens, the generator is only capable of generating examples close to these modes 
and fails to generate a high variety from different prototypes available in the training data.

In this paper, we propose \emph{Mixture Density GAN} that is capable of generating high-quality samples, and in addition copes with mode collapse problem and enables the GAN to generate samples with a high variety.
The central idea of \textbf{Mixture Density GAN (MD-GAN)} is to enable the discriminator to create several clusters in its output embedding space for real images, and therefore provides better means to distinguish not only real and fake, but also between different kinds of real images.
The discriminator in MD-GAN forms a number of clusters\footnote{The number of clusters is a parameter that can be set.} with embeddings of real images which represent clusters in the real data. To fool the discriminator, the generator then has to generate images that discriminator has to embed close to the center of these clusters. As there are multiple clusters, the generator can discover various modes by generating images that end up in various clusters.

MD-GAN's Discriminator uses a $d$-dimensional embedding space and is provided with an objective function that pushes it towards forming clusters in this space which are arranged in the form of a \emph{Simplex}\footnote{A simplex is defined as a generalization of the notion of a tetrahedron with $d$ dimensions and $d+1$ vertices\protect\cite{montash907simplex}.}: each cluster center is located in one of the vertices of this simplex.
In our empirical experiments, we use four benchmark datasets of images and one synthetic dataset to demonstrate the ability of MD-GAN to generate samples with good quality and high variety, and to avoid mode collapse.
Comparing our results to state-of-the-art methods in terms of the Fr\'echet Inception Distance (FID)~\cite{heusel2017gans} as well as the number of discovered modes, 
we will demonstrate that MD-GAN can achieve state-of-the-art level results.

\section{Mixture Density Generative Adversarial Networks}
\label{sec:simplexgan}

\subsection{Mixture Density GAN: The Intuition}
\label{subsec:intuition}

\begin{figure} 
  \centering
  \fbox{\includegraphics[width=.5\linewidth]{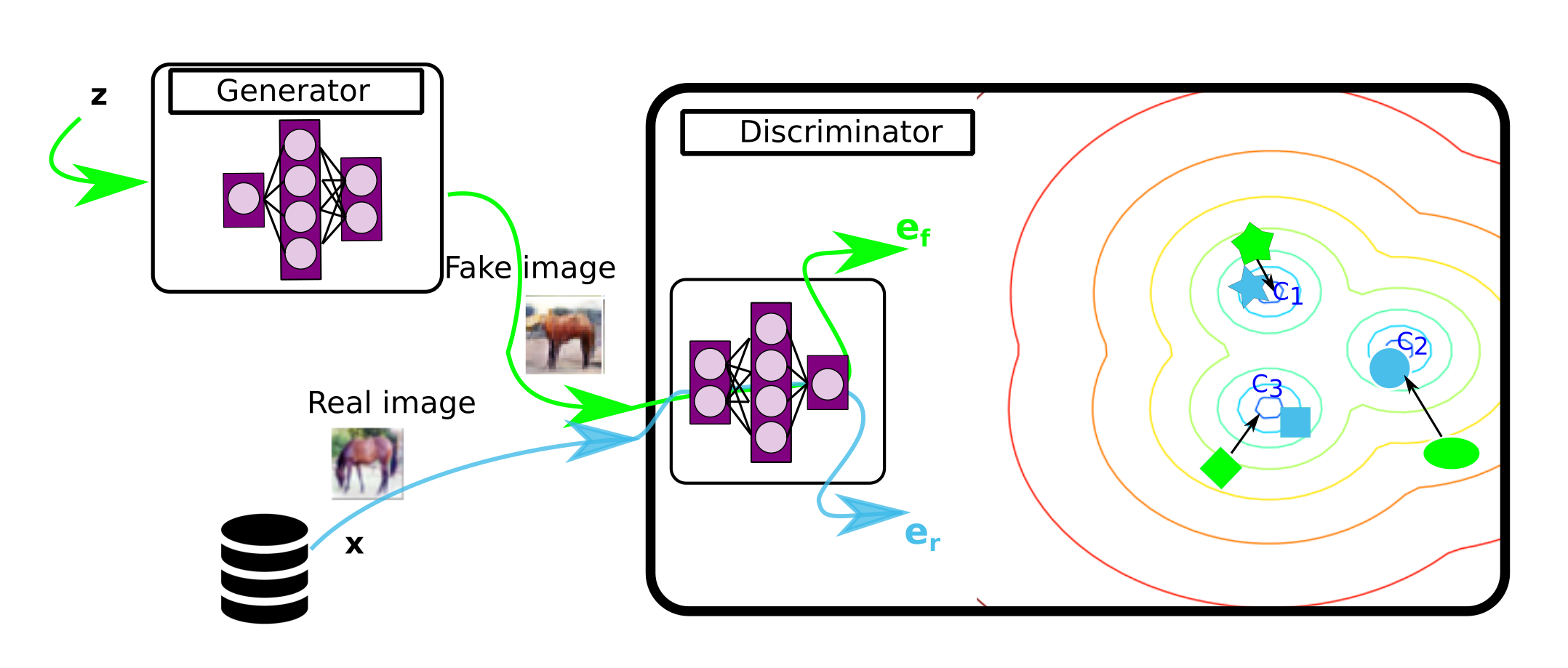}}
  \caption{Block diagram of Mixture Density GAN. This figure should be viewed in colour.}
 \label{fig:block_diagram}
\end{figure}

As explained in the introduction, the basic idea in Mixture Density GAN is to encourage the discriminator to form a number of clusters from embeddings of real images.
As also mentioned above, these clusters will be positioned in an equi-distant way, their center vectors forming a \emph{simplex}.
Each cluster is represented by a Gaussian kernel, the whole collection
thus makes up a \emph{mixture of Gaussians}, which we will call
a \emph{Simplex Gaussian Mixture Model (SGMM)}.
Each of the clusters draw embeddings of fake images towards their center.
This is achieved by using the SGMM as a likelihood function.
Each Gaussian kernel spreads its density over the embedding space: the closer an embedding to the center of a cluster, the more density it gets and therefore, the more likelihood reward it receives.

By defining a likelihood function with the parameters of a SGMM, 
in each update we train the discriminator to encode real images to the centers of the clusters. The resulting SGMM creates a mixture of clusters that draws the real embeddings towards the cluster centers (see Figure~\ref{fig:block_diagram}).
Because of the multiple clusters, the generator will be rewarded if it generates samples that end up in any of these clusters.
Thus, if the fake embeddings are well spread around the clusters -- which they are likely to be at the beginning of training, when they are essentially just random projections --, it is likely that most of the clusters will draw the fake embeddings.
Therefore, the generator will tend to learn to
generate samples with more variety to cover all of the clusters, which ideally results in discovering the modes present in the data and directly addresses the mode collapse problem.
On the other hand, it is reasonable to expect the discriminator to create such clusters based on relevant similarities in the data, since it is trained as a classifier and therefore needs to learn a meaningful distance in its embedding space.
Experiments described in results Section will confirm our intuition.

\subsection{Mixture Density GAN: The Model}
\label{subsec:simplegan}

As in the vanilla GAN, MD-GAN consists of a generator $G$ and a discriminator $D$.
MD-GAN uses a mixture of Gaussians in its objective functions whose mean vectors are placed in the cartesian coordinates of the vertices in a $d$-dimensional simplex. 

\textbf{Discriminator}: The discriminator $D$ in MD-GAN is a neural network  with $d$-dimensional output. 
For an input image $\boldsymbol{x}$, the discriminator creates an embedding $\boldsymbol{e}$ which is simply the activations of the last layer of the network for input $\boldsymbol{x}$.

The SGMM in MD-GAN is a Gaussian mixture with the following properties:
1) The individual components are $d$-dimensional multivariate Gaussians (where $d$ is the output/embedding dimensionality of the discriminator network).
2) The model comprises $d+1$ Gaussian components, whose mean vectors are exactly the coordinates of the vertices of a simplex.
3) The covariance matrices are diagonal and have equal values on the main diagonal, in all the components. Thus, all components are spherical Gaussians.
4) The component weights $w_i$ are either $1$ 
if the component produces the highest likelihood among all components, or $0$ otherwise.

For an embedding $e$ produced by the discriminator $D$, we define the following \emph{likelihood function:}

\begin{equation}
\label{eq:lk}
lk(\boldsymbol{e}) = \sum_{i=1}^{n} w_{i} \cdot \Phi \Big( \boldsymbol{e}; \boldsymbol{\mu}_{i}, \boldsymbol{\Sigma}_{i} \Big)
\end{equation}
where $\Phi$ is the Gaussian PDF, $w_i$ is the mixture weight, $\boldsymbol{\mu}_i$ is the mean vector, and $\boldsymbol{\Sigma}_i$ is the covariance matrix for Gaussian component $i$.

When a discrimination between real and fake images is needed, the discriminator first encodes the input image $\boldsymbol{x}$ into the embedding $\boldsymbol{e}$.
Then, a likelihood $lk(\boldsymbol{e})$ is calculated for this embedding.
$lk(\boldsymbol{e})$ will be interpreted as the probability of $\boldsymbol{e}$ being an embedding of a real image, given the current model. \\

\textbf{Generator}: The generator $G$ in MD-GAN is a regular neural network decoder, decoding a random noise $z$ from a random distribution $P_{\mathit{z}}$ into an image.

\subsection{The Mixture Density GAN Objectives}
\label{subsec:simplegan_objective}

Denoting the encoding (output of the encoder, also referred to as the embedding) of an image $\boldsymbol{x}$ by discriminator D as D$(\boldsymbol{x})$, 
we propose the MD-GAN's objectives as follows:

\begin{align}\label{eq:simplexgan_disc_minimax}
\begin{split}
\min_{G} &\max_{D} \mathcal{L}(G,D)= \\
&\min_{G} \max_{D} \mathbb{E}_{\boldsymbol{x} \sim p_{\mathit{data}} } \big[ \log( lk(D(\boldsymbol{x}))) \big]\\
&+\mathbb{E}_{\boldsymbol{z}\sim p_{\boldsymbol{z}} } \big[  \log( \lambda - lk(D(G(\boldsymbol{z})))) \big]
\end{split}
\end{align}
where the likelihood lk($\boldsymbol{e}$) for the given image embedding $\boldsymbol{e} = D(\boldsymbol{x})$ is as defined in Eq.\eqref{eq:lk}.
We set $\lambda$ to be the maximum value of the likelihood function $lk$ in order to have only positive values in the logarithm in Eq.\eqref{eq:simplexgan_disc_minimax} (see also Experimental Setup Section.)

\section{Empirical Results}
\label{sec:empirical_results}

\begin{figure}[H]
\centering
\begin{subfigure}{.18\textwidth}
  \centering
  \includegraphics[width=1.\linewidth]{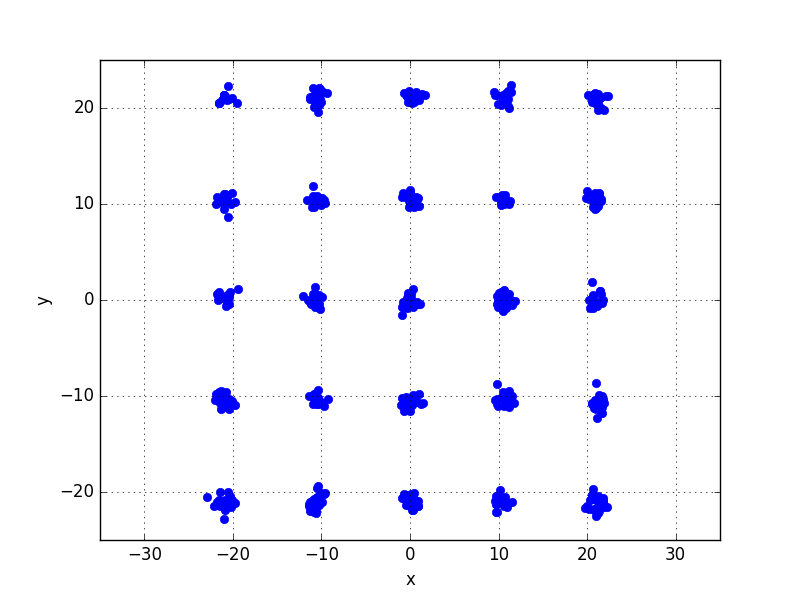}
  \caption{Real}
  \label{fig:img_grid_real}
\end{subfigure}%
\begin{subfigure}{.18\textwidth}
  \centering
  \includegraphics[width=1.\linewidth]{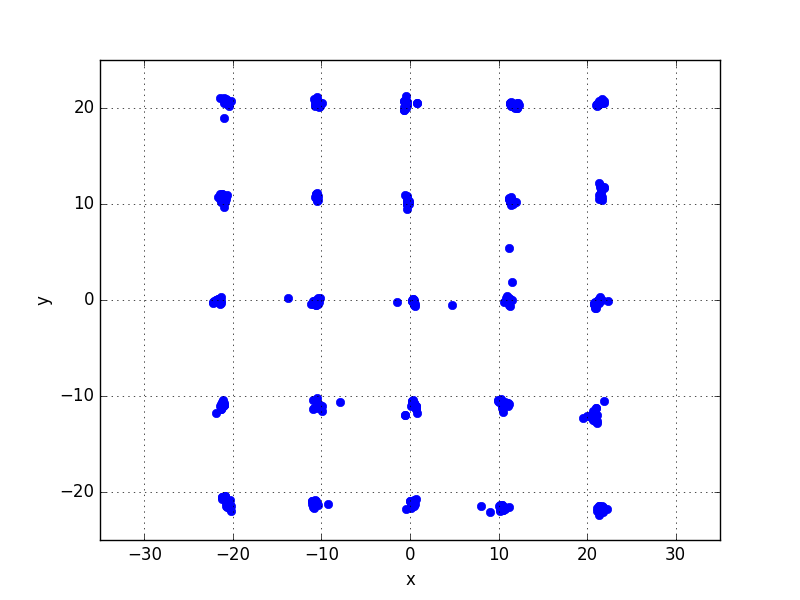}
  \caption{MD-GAN}
  \label{fig:img_grid_sgan}
\end{subfigure}%
\begin{subfigure}{.18\textwidth}
  \centering
  \includegraphics[width=1.\linewidth]{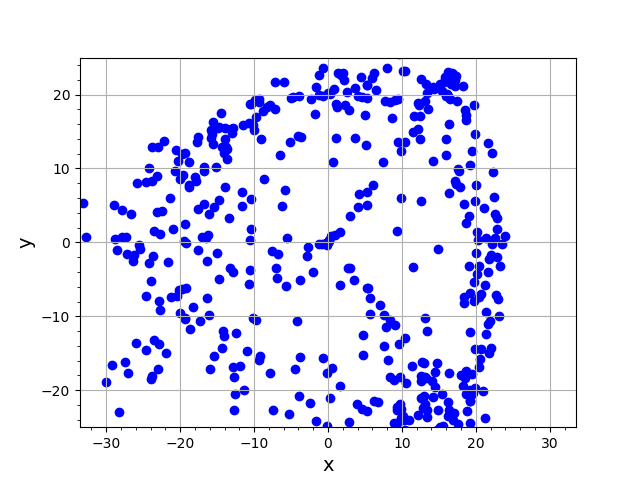}
  \caption{WGAN}
  \label{fig:img_grid_wgan}
\end{subfigure}
\begin{subfigure}{.18\textwidth}
  \centering
  \includegraphics[width=1.\linewidth]{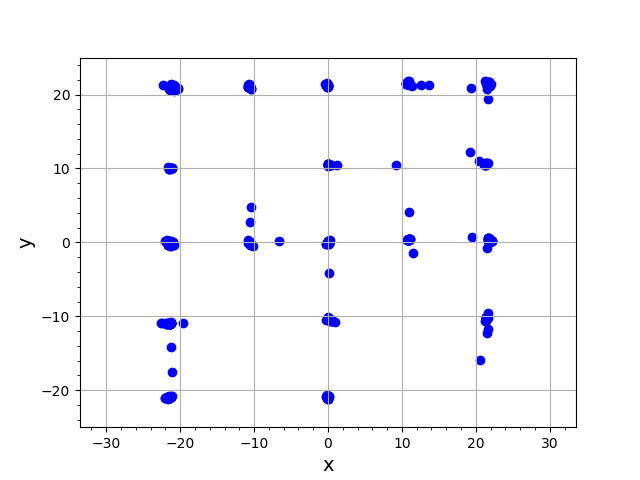}
  \caption{Vanil}
  \label{fig:img_grid_vanillagan}
\end{subfigure}
\begin{subfigure}{.18\textwidth}
  \centering
  \includegraphics[width=1.\linewidth]{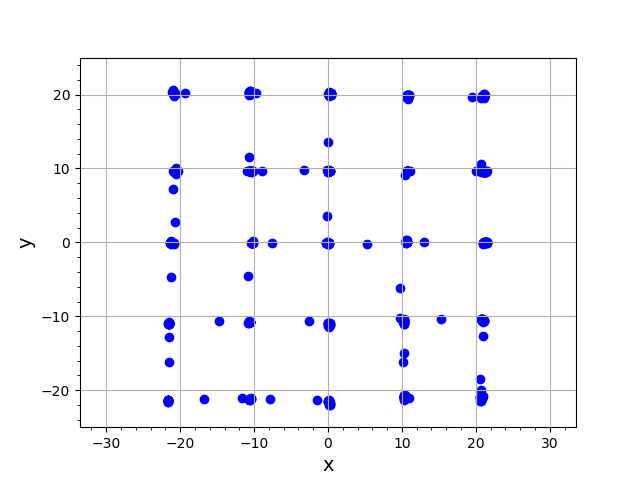}
  \caption{GGAN}
  \label{fig:img_grid_geogan}
\end{subfigure}
\caption{a-e) Examples generated on our synthetic dataset via different methods. 
Results of c-d are taken from~\cite{LimY17geo}.}
\label{fig:synthetic_grid}
\end{figure}

The results of mode-collapse experiments are provided in Figure~\ref{fig:synthetic_grid} and the experimental results evaluated with FID can be found in Table~\ref{tab:fid_results}. These results are obtained by using only a simple DCGAN~\cite{radford2015unsupervised}architectures with limited number of parameters that are usually used to demonstrate the abilities of an objective in achieving high-quality image generation.
As can be seen, MD-GAN discovers all modes in the synthetic dataset and achieves the lowest (best) FID among all baselines.

\begin{table}
\centering
\caption{On the left: FIDs on different datasets from different methods.
*) Results from\protect\cite{lucic2017gans}
which are best FIDs obtained in a large-scale hyper-parameter search for each data set.
Lower FID values represent higher quality for generated images.
On the right: Results of mode collapse experiments on 2D-grid of 25 Gaussians.
All the results are the average of 5 runs.
}
\label{tab:fid_results}
\begin{minipage}{.52\linewidth}
\centering
\begin{tabular}{{l|c|c|c|c|c|c}}
\hline\noalign{\smallskip}
\multicolumn{3}{c|}{} &    MNT &   FMNT & CFR&Clb\\
\noalign{\smallskip}\hline\noalign{\smallskip}
\multicolumn{3}{c|}{Real *}& 1.2   &   2.6  &   5.1  & 2.2 \\ 
\hline
\multicolumn{3}{c|}{Vanilla~\cite{goodfellow2014generative}}&  6.7   &   26.6  &  58.6  &  58.0 \\   
\multicolumn{3}{c|}{WGAN~\cite{arjovsky2017wasserstein}} &  6.8   &   18.0  &   55.9  &   42.9 \\   
\multicolumn{3}{c|}{WGAN-GP~\cite{gulrajani2017improved}}&  8.9   &   20.6  &  52.9  &   26.8  \\ 
\multicolumn{3}{c|}{DRAGAN~\cite{kodali2017convergence}} &  7.7   &  26.0   & 68.5  & 41.4\\ 
\multicolumn{3}{c|}{BEGAN~\cite{berthelot2017began}} & 12.3    &   33.2   &  71.4 & 38.1\\
\multicolumn{3}{c|}{MD-GAN} & \textbf{6.29}  &   \textbf{11.79}  &\textbf{36.80} & \textbf{24.51} \\
\noalign{\smallskip}\hline
\end{tabular}
 \end{minipage}%
\begin{minipage}{.52\linewidth}
\centering
\begin{tabular}{{c|c|c}}
\hline\noalign{\smallskip}
 &modes&\% hq\\
\hline\noalign{\smallskip}
Vanilla~\cite{goodfellow2014generative}&  3.3    &0.5   \\ 
ALI~\cite{dumoulin2016adversarially} &     15.8  &     1.6\\ 
Unrol.~\cite{metz2016unrolled}&  23.6 & 16  \\ 
VEEGAN~\cite{srivastava2017veegan}&  24.6 &    40\\ 
\textbf{MD-GAN}&$\mathbf{25}$& $\mathbf{99.36\rpm2.28}$\\\hline
\end{tabular}
\end{minipage}
\end{table}

\section{Conclusion}
\label{sec:conclusion}

In this paper, we proposed a novel GAN variant called
Mixture Density GAN, which succeeds in generating high-quality images and in addition alleviates the mode collapse problem by allowing
the Discriminator to form separable clusters
in its embedding space, which in turn leads the Generator to
generate data with more variety.
We analysed the optimum discriminator 
and showed that it is achieved when the generated
and the real distribution match exactly and further discussed the relations to the vanilla GAN.
We demonstrated the ability of Mixture Density GAN to deal with mode collapse and generate realistic images
using a synthetic dataset and 4 image benchmarks.

\bibliographystyle{plain}
\bibliography{main_file}
\end{document}